\documentclass[runningheads]{llncs}
\usepackage{booktabs}
\usepackage{multirow}
\usepackage{multicol}
\usepackage{bm}
\usepackage{colortbl}
\usepackage{float}
\usepackage{xcolor}
 
\makeatletter
\newcommand{\thickhline}{%
    \noalign {\ifnum 0=`}\fi \hrule height 2pt
    \futurelet \reserved@a \@xhline
}
\definecolor{mygray}{gray}{.9}

\usepackage{graphicx}
\usepackage{comment}
\usepackage{amsmath,amssymb} % define this before the line numbering.
\usepackage{color}
\usepackage{marvosym}
\usepackage{lipsum}

\usepackage[,pagebackref,breaklinks,colorlinks]{hyperref}

\usepackage[accsupp]{axessibility}  % Improves PDF readability for those with disabilities.

\usepackage[capitalize]{cleveref}
\crefname{section}{Sec.}{Secs.}
\Crefname{section}{Section}{Sections}
\Crefname{table}{Table}{Tables}
\crefname{table}{Tab.}{Tabs.}

\newcommand{\gray}[1]{\textcolor{gray}{#1}}
\newlength\savewidth\newcommand\shline{\noalign{\global\savewidth\arrayrulewidth
  \global\arrayrulewidth 1pt}\hline\noalign{\global\arrayrulewidth\savewidth}}
\newcommand{\tablestyle}[2]{\setlength{\tabcolsep}{#1}\renewcommand{\arraystretch}{#2}\centering\footnotesize}

\newcommand\blfootnote[1]{%
\begingroup
\renewcommand\thefootnote{}\footnote{#1}%
\addtocounter{footnote}{-1}%
\endgroup
}

\begin{document}
% \linenumbers
\pagestyle{headings}
\mainmatter
\def\ECCVSubNumber{7175}  % Insert your submission number here

\title{ActiveNeRF: Learning where to See with Uncertainty Estimation} % Replace with your title

%\begin{comment}
% \titlerunning{ECCV-22 submission ID \ECCVSubNumber} 
% \authorrunning{ECCV-22 submission ID \ECCVSubNumber} 
% \author{Anonymous ECCV submission}
% \institute{Paper ID \ECCVSubNumber}
%\end{comment}
%******************

% \begin{comment}
\titlerunning{ActiveNeRF}
% \author{Xuran Pan\inst{1} \orcidlink{0000-0002-6434-8159} \and
% Zihang Lai\inst{2}\textsuperscript{\dag} \orcidlink{0000-0002-9872-0756} \and
% Shiji Song\inst{1}\orcidlink{0000-0001-7361-9283} \and
% Gao Huang\inst{1}\textsuperscript{\Letter}\orcidlink{0000-0002-7251-0988}}
\author{Xuran Pan\inst{1}  \and
Zihang Lai\inst{2}\textsuperscript{\dag}  \and
Shiji Song\inst{1} \and
Gao Huang\inst{1}\textsuperscript{\Letter}}

% \orcidID{2222--3333-4444-5555}
%
\authorrunning{X. Pan et al.}
\institute{Tsinghua University, Beijing 100084, China\\
\email{pxr18@mails.tsinghua.edu.cn} \email{\{shijis, gaohuang\}@tsinghua.edu.cn}\and
Carnegie Mellon University, Pennsylvania 15213, United States
\email{zihangl@andrew.cmu.edu}\blfootnote{\dag \ Work done during an internship at Tsinghua University.}\blfootnote{\Letter \ Corresponding author.}}
% \end{comment}
%******************
\maketitle

\begin{abstract}
% The abstract should summarize the contents of the paper. LNCS guidelines
% indicate it should be at least 70 and at most 150 words. It should be set in 9-point
% font size and should be inset 1.0~cm from the right and left margins.
% \dots
Recently, Neural Radiance Fields (NeRF) has shown promising performances on reconstructing 3D scenes and synthesizing novel views from a sparse set of 2D images. Albeit effective, the performance of NeRF is highly influenced by the quality of training samples. With limited posed images from the scene, NeRF fails to generalize well to novel views and may collapse to trivial solutions in unobserved regions. This makes NeRF impractical under resource-constrained scenarios. In this paper, we present a novel learning framework, \textit{ActiveNeRF}, aiming to model a 3D scene with a constrained input budget. Specifically, we first incorporate uncertainty estimation into a NeRF model, which ensures robustness under few observations and provides an interpretation of how NeRF understands the scene. On this basis, we propose to supplement the existing training set with newly captured samples based on an active learning scheme. By evaluating the reduction of uncertainty given new inputs, we select the samples that bring the most information gain. In this way, the quality of novel view synthesis can be improved with minimal additional resources. Extensive experiments validate the performance of our model on both realistic and synthetic scenes, especially with scarcer training data. Code will be released at \url{https://github.com/LeapLabTHU/ActiveNeRF}.

\end{abstract}

\begin{figure}
    \centering
    \includegraphics[width=0.9\linewidth]{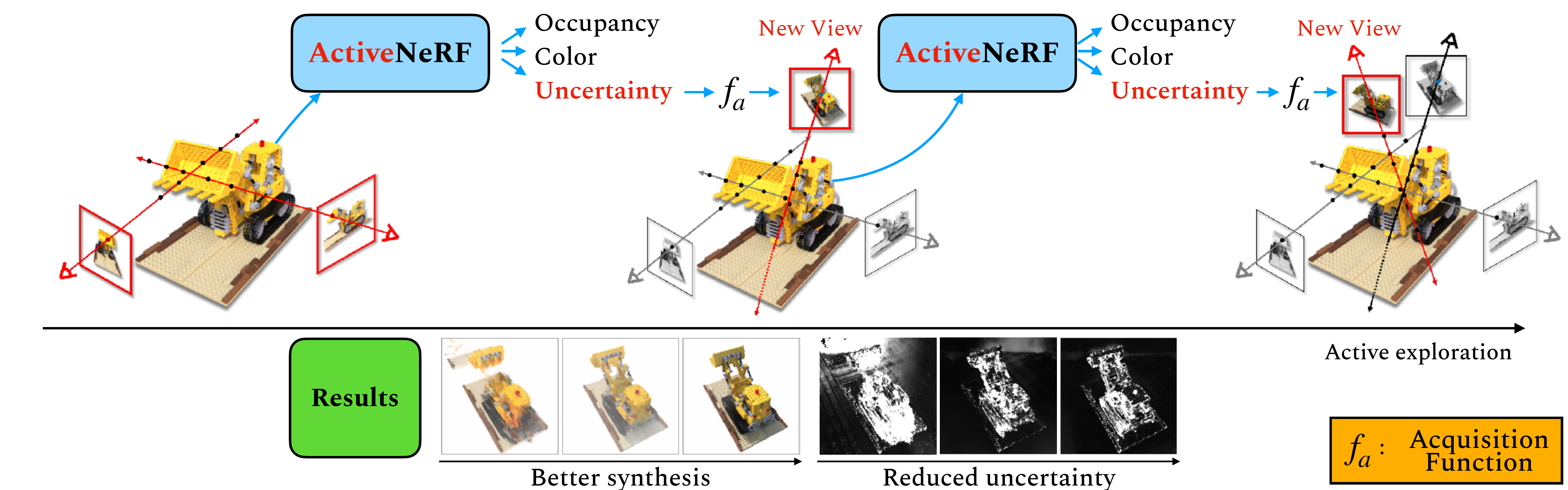}
    \vskip -0.1in
    \caption{\textbf{ActiveNeRF:} We present a flexible learning framework that \emph{actively} expands the existing training set with newly captured samples based on an Active Learning scheme. ActiveNeRF incorporates uncertainty estimation into a NeRF model and evaluates the reduction of scene uncertainty at unobserved novel views. By selecting the view that brings the most information gain, the quality of novel view synthesis can be improved with minimal additional resources.}
    \label{figure1}
\end{figure}
\vskip -0.1in

\section{Introduction}
The task of synthesizing novel views of a scene from a sparse set of images has earned broad research interest in recent years. With the advent of neural rendering techniques, Neural Radiance Fields (NeRF) \cite{mildenhall2020nerf} shows its potential on rendering photo-realistic images and inspires a new line of research \cite{zhang2020nerf++,park2020deformable,pumarola2021d}. Different from traditional Structure-from-Motion \cite{andrew2001multiple} or image-based rendering \cite{shum2008image} approaches, NeRF models the emitted radiance values and volume densities in a 3D scene as a function of continuous 5D coordinates, including spatial locations $x,y,z$ and viewing directions $\theta, \phi$. The learned implicit function expresses a compact representation of the scene and enables free-viewpoint synthesis through volume rendering.

\begin{figure}[t]
    \centering
    \includegraphics[width=0.85\linewidth]{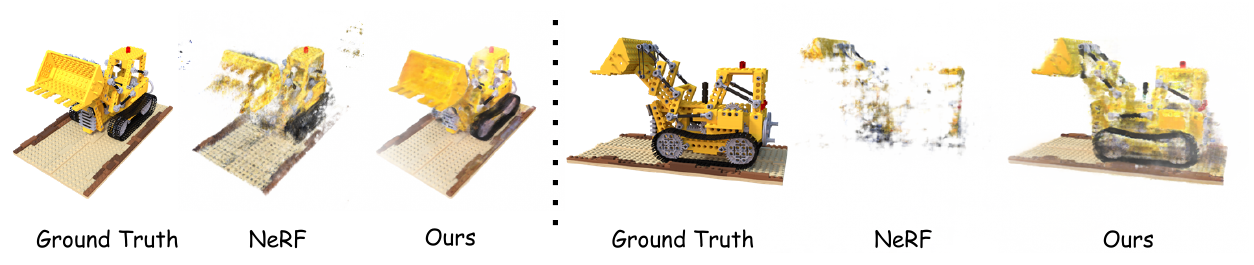}
    \caption{\textbf{Novel view synthesis of NeRF with partial observations.} The models are trained with 10 posed images where observations from the left side are removed from the training set. While our model can still generate reasonably good synthesis results, the original NeRF shows large errors or completely fails to generate meaningful content.}
    \label{fig:fig3}
\end{figure}

Despite its success in synthesizing high-quality images, the learning scheme for a NeRF model puts forward higher demands on the training data. First, NeRF usually requires a large number of posed images and is proved to generalize poorly with limited inputs~\cite{yu2021pixelnerf}. Second, it takes a whole observation in the scene to train a well-generalized NeRF. As illustrated in Figure \ref{fig:fig3}, if we remove observations of a particular part in the scene, NeRF fails to model the region and tends to collapse (\textit{e.g.,} predicting zero density everywhere in the scene) instead of performing reasonable predictions. This poses challenges under real-world applications such as robot localization and mapping, where capturing training data can be costly, and perception of the entire scene is required \cite{paull2016unified,khosoussi2019reliable,torres2017robot}.

In this paper, we focus on the context with constrained input image budget and attempt to address these limitations by leveraging the training data in the most efficient manner. As shown in Figure \ref{figure1}, we first introduce uncertainty estimation into the NeRF framework by modeling the radiance values of each location as a Gaussian distribution. This imposes the model to provide larger variances in the unobserved region instead of collapsing to a trivial solution. On this basis, we resort to the inspiration from active learning and propose to capture the most informative inputs as supplementary to the current training data. Specifically, given a hypothetical new input, we analyze the posterior distribution of the whole scene through Bayesian estimation, and use the subtraction of the variance from prior to posterior distribution as the information gain. This finally serves as the criterion for capturing new inputs, and thus raises the quality of synthesized views with minimal additional resources. Extensive experiments show that NeRF with uncertainty estimation achieves better performances on novel view synthesis, especially with scarce training data. Our proposed framework based on active learning, dubbed \textbf{ActiveNeRF}, also shows superior performances on both synthetic and realistic scenes, and outperforms several heuristic baselines.

\section{Related Works}
\subsection{Novel View Synthesis}
Synthesizing novel views of a 3D scene from a sparse set of 2D images is a long-standing problem in computer vision. Earlier work, including Structure-from-Motion \cite{andrew2001multiple} or image-based rendering \cite{shum2008image}, mostly reconstruct a scene in sparse representations. On this basis, bundle adjustment \cite{triggs1999bundle} and lighting-based approaches \cite{levoy1996light} consider the light and reluctance properties to synthesize photo-realistic images. More recently, the neural rendering technique has been introduced to the scene representation task, which inspires a line of research to model the 3D scene as a continuous representation. Scene Representation Network (SRN) \cite{sitzmann2019scene} first models the scene as a function of 3D coordinates, which are then used to predict the intersections of object surfaces and the corresponding emitted color. Following SRN, Neural Radiance Fields (NeRF) \cite{mildenhall2020nerf} considers the volume density and view-dependent emitted color in the scene and models with a simple but effective multi-layer perceptron. The outputs in each location of the scene are combined with neural rendering techniques to synthesize novel views.

Many researches follow the step of NeRF and extend the original framework from different perspectives \cite{wang2021nerf,arandjelovic2021nerf,kosiorek2021nerf}. NeRF++ \cite{zhang2020nerf++} analyzes the modeling capacity of NeRF and proposes an inverted sphere parameterization approach to model unbounded 3D scenes. FastNeRF \cite{garbin2021fastnerf} accelerates rendering procedure in NeRF to achieve real-time view synthesis. D-NeRF \cite{pumarola2021d} and other related approaches propose to model dynamic scenes with moving objects. NeRF-W \cite{martin2021nerf}, on the other hand, focuses on modeling the transient objects varying from different images. Several works further extend NeRF-based models to represent scenes conditioned on a scene prior, which enables NeRF to generalize to new scenes. 

More related to our work, several researches have also addressed the problem of NeRF under the limited input setting. Pixel-NeRF \cite{yu2021pixelnerf} proposes to encode the image-level features into the radiance field and trains a NeRF model that can generalize across the scene. MVSNeRF \cite{chen2021mvsnerf} applies 3D CNN to reconstruct a neural encoding volume with per-voxel neural features. GRF \cite{trevithick2021grf} back-projects points to input images and gathers per-pixel features from each view. These approaches incorporate image features into original coordinate-based embeddings. DietNeRF \cite{jain2021putting} introduces additional semantic consistency loss with pretrained CLIP \cite{radford2021learning} models. Compared to these works, we are the first to address the limitation of NeRF from the data perspective and effectively increase the upper bound of the model with minimum additional resources. Also, the uncertainty estimation module in our framework is orthogonal to these approaches and can serve as a plug-and-play module to further boost their performances.

\subsection{Uncertainty Estimation}
The computer vision community has seen the value of uncertainty estimation in various research fields. Measuring the uncertainty of a neural network can both enhance the interpretability of the model outputs and reduce the risk of making critical faults. Based on the Bayesian rule, several approaches formulate uncertainty as a probability distribution over either the model parameters or model outputs. Bayesian Neural Networks (BNN) \cite{mackay1995bayesian,kononenko1989bayesian} approaches measure the uncertainty as posterior distribution, which usually require approximate inference methods, \textit{e.g.,} variational inference. Dropout variational inference \cite{gal2016dropout,kingma2015variational} estimates the model uncertainty with dropout layers in the network by performing multiple inferences for the same input. 

Early research has also explored the possibility of applying uncertainty estimation in the field of novel view synthesis. NeRF-W \cite{martin2021nerf} introduces uncertainty to model the transient objects in the scene. Compared to our approach, the uncertainty estimation in NeRF-W focuses on the differences across the images rather than the inherent noise inside the training data. Another concurrent work S-NeRF \cite{shen2021stochastic} models the uncertainty of the scene with variational inference. Although the uncertainty correlates well with the predictive error, S-NeRF performs qualitatively worse (\textit{e.g.,} it shows blurry edges in the synthesis results) than the original model. It also requires multiple identical inferences to obtain the uncertainty map. Compared to these two approaches, our simple yet effective uncertainty estimation framework strictly follows the volume rendering procedure, and shows on par or better performances over the original NeRF model under various training data settings. The proposed uncertainty modeling is also a necessary component of the full ActiveNeRF framework: the uncertainty estimation serves as the basis to evaluate the new images.

\subsection{Active Learning}
Active learning has been widely studied in various computer vision tasks, including image classification \cite{fu2018scalable}, image captioning \cite{miller2014adversarial}, and object detection \cite{bengar2019temporal}. Active learning can be categorized into two classes: representativeness-based and informativeness-based approaches. Representativeness methods rely on selecting examples by increasing the diversity of the training set. Core-set technique \cite{sener2017active} selects the samples by evaluating the Euclidean distance between candidates and labeled samples in the feature space. Also, several researches resort to the techniques in adversarial training \cite{sinha2019variational} or self-supervised training \cite{bengar2021reducing}, and select samples with an additional network, \textit{e.g.,} a discriminator. More related to our work, informativeness methods measure the uncertainty of each data and select the most uncertain ones from an unlabelled data pool. With the uncertainty estimation approaches in the previous section, the selection criterion can be used in both Bayesian \cite{gal2017deep} and non-Bayesian \cite{li2013adaptive} frameworks.

To the best of our knowledge, ActiveNeRF is the first approach to incorporating active learning scheme into the NeRF optimization pipeline. Unlike other works that focus on improving model capacities, we analyze the inherent imperfection of the training data, thereby increasing the synthesis quality of NeRF models with higher data efficiency. This is crucial for resource-constrained scenarios in real-world applications. 
\section{Background}
\label{background}
In this section, we first briefly review the Neural Radiance Fields (NeRF) framework and introduce some implementation details. 

NeRF models a scene as a continuous function $F_{\theta}$ which outputs emitted radiance value and volume density. Specifically, given a 3D position $\text{x}\!=\!(x,y,z)$ in the scene and a viewing direction parameterized as a 3D Cartesian unit vector $\text{d}\!=\!(d_x,d_y,d_z)$, a multi-layer perceptron model is adopted to produce the corresponding volume density $\sigma$ and color $\text{c}=(r,g,b)$ as follows:
\begin{align}
\setlength{\abovedisplayskip}{1ex}
\label{mlp1}
    [\sigma, f] = &\text{MLP}_{\theta_1}(\gamma_{\text{x}}(\text{x})), \\
    c = &\text{MLP}_{\theta_2}(f, \gamma_\text{d}(\text{d})),
\setlength{\belowdisplayskip}{1ex}
\end{align}
where $\gamma_{\text{x}}(\cdot)$ and $\gamma_\text{d}(\cdot)$ are the positional encoding functions, and $f$ represents the intermediate feature independent from viewing direction $\text{d}$. An interesting observation is that the radiance color is only affected by its own 3D coordinates and the viewing direction, which makes it independent from other locations.

To achieve free view synthesis, NeRF renders the color of rays passing through the scene with the volume rendering technique. Let $\text{r}(t)=\text{o}+t\text{d}$ be a camera ray with camera center $\text{o}\in \mathcal{R}^3$ through a given pixel on the image plane, the color of the pixel can be formulated as:
\begin{equation}
\setlength{\abovedisplayskip}{1ex}
    C(\text{r})=\int_{t_n}^{t_f}T(t)\sigma(\text{r}(t))\text{c}(\text{r}(t),\text{d})dt,
\setlength{\belowdisplayskip}{1ex}
\end{equation}
where $T(t)\!=\!\text{exp}(\!-\!\int_{t_n}^t\sigma(\text{r}(s))ds)$ denotes the accumulated transmittance, and $t_n$ and $t_f$ are the near and far bounds in the scene. To make the rendering process tractable, NeRF approximates the integral based on stratified sampling, and formulates it as a linear combination of sampled points:
\begin{equation}
\setlength{\abovedisplayskip}{1ex}
\label{render}
    \hat{C}(\text{r})=\sum_{i=1}^{N_s}\alpha_ic(\text{r}(t_i)), \ \alpha_i\!=\!\text{exp}(-\sum_{j=1}^{i-1}\sigma_j\delta_j)(1-\text{exp}(-\sigma_i\delta_i)),
\setlength{\belowdisplayskip}{1ex}
\end{equation}
where $\delta_i=t_{i+1}-t_i$ is the distance between adjacent samples, and $N_s$ denotes the number of samples. On this basis, NeRF optimizes the continuous function $F_{\theta}$ by minimizing the squared reconstruction errors between the ground truth from RGB images $\{\mathcal{I}_{i=1}^N\}$, and the rendered pixel colors.

To improve the sampling efficiency, NeRF optimizes two parallel networks simultaneously, and denote them as coarse and fine models respectively. The sampling strategy for the fine model is improved according to the result of the coarse model, where the samples are biased towards more relevant parts. In all, the optimization loss is parameterized as:
\begin{equation}
\setlength{\abovedisplayskip}{1ex}
    \sum_{i}\|C(\text{r}_i)-\hat{C}^c(\text{r}_i)\|_2^2+\|C(\text{r}_i)-\hat{C}^f(\text{r}_i)\|_2^2,
\setlength{\belowdisplayskip}{1ex}
\end{equation}
where $\text{r}_i$ is sampled ray, and $C(\text{r}_i), \hat{C}^c(\text{r}_i), \hat{C}^f(\text{r}_i)$ correspond to the ground truth, coarse model prediction, and fine model prediction respectively.

\section{NeRF with Uncertainty Estimation}
\label{uncert_est}
In this paper, we focus on the context in some real-world applications, where the number of training data is within a limited budget. It has been proved in existing research~\cite{yu2021pixelnerf} that NeRF fails to generalize well from one or few input views. If with incomplete scene observation, the original NeRF framework tends to collapse to trivial solutions by predicting the volume density as 0 for the unobserved regions. 

As a remedy, we propose to model the emitted radiance value of each location in the scene as a Gaussian distribution instead of a single value. The predicted variance can serve as the reflection of the aleatoric uncertainty concerning a certain location. Through this, the model is imposed to provide larger variances in the unobserved region instead of collapsing to the trivial solution. 

Specifically, we define the radiance color of a location $\text{r}(t)$ follows a Gaussian distribution parameterized by mean $\bar{c}(\text{r}(t))$ and variance $\bar{\beta}^2(\text{r}(t))$. Following previous researches in Bayesian neural networks, we take the model output as the mean, and add an additional branch to the MLP network in Eq.(\ref{mlp1}) to model the variance as follows:
\begin{align}
\setlength{\abovedisplayskip}{1ex}
    [\sigma, f, \beta^2(\text{r}(t))] = &\text{MLP}_{\theta_1,\theta_3}(\gamma_\text{x}(\text{r}(t))), \\
    \bar{c}(\text{r}(t)) = &\text{MLP}_{\theta_2}(f, \gamma_\text{d}(\text{d})).
\setlength{\belowdisplayskip}{1ex}
\end{align}
Softplus function is further adopted to produce a validate variance value:
\begin{equation}
\setlength{\abovedisplayskip}{1ex}
    \bar{\beta}^2(\text{r}(t)) = \beta_0^2 + \text{log}(1+\text{exp}(\beta^2(\text{r}(t))),
\setlength{\belowdisplayskip}{1ex}
\end{equation}
where $\beta_0^2$ ensures a minimum variance for all the locations. 

In the rendering process, the new neural radiance field with uncertainty can be similarly performed through volume rendering. As we have mentioned in Sec. \ref{background}, the design paradigm in the NeRF framework provides two valuable prerequisites. (1) The radiance color of a particular position is only affected by its own 3D coordinates, which makes the distribution of different positions independent from each other. (2) Volume rendering can be approximated as linear combination of sampled points along the ray. On this basis, if we denote the Gaussian distribution of a position at $\text{r}(t)$ as $c(\text{r}(t))\sim \mathcal{N}(\bar{c}(\text{r}(t)), \bar{\beta}^2(\text{r}(t)))$, the rendered value along this ray naturally follows Gaussian distribution:
\begin{equation}
\label{render_uct}
\setlength{\abovedisplayskip}{1ex}
    \hat{C}(\text{r}) \sim \mathcal{N}(\sum_{i=1}^{N_s}\alpha_i\bar{c}(\text{r}(t_i)), \sum_{i=1}^{N_s}\alpha_i^2\bar{\beta}^2(\text{r}(t_i)))\sim \mathcal{N}(\bar{C}(\text{r}), \bar{\beta}^2(\text{r})),
\setlength{\belowdisplayskip}{1ex}
\end{equation}
where the $\alpha_i$s are the same as in Eq.(\ref{render}), and $\bar{C}(\text{r}), \bar{\beta}^2(\text{r})$ denote the mean and variance of the rendered color through the sampled ray $\text{r}$.

\begin{figure}[t]
    \centering
    \includegraphics[width=0.85\linewidth]{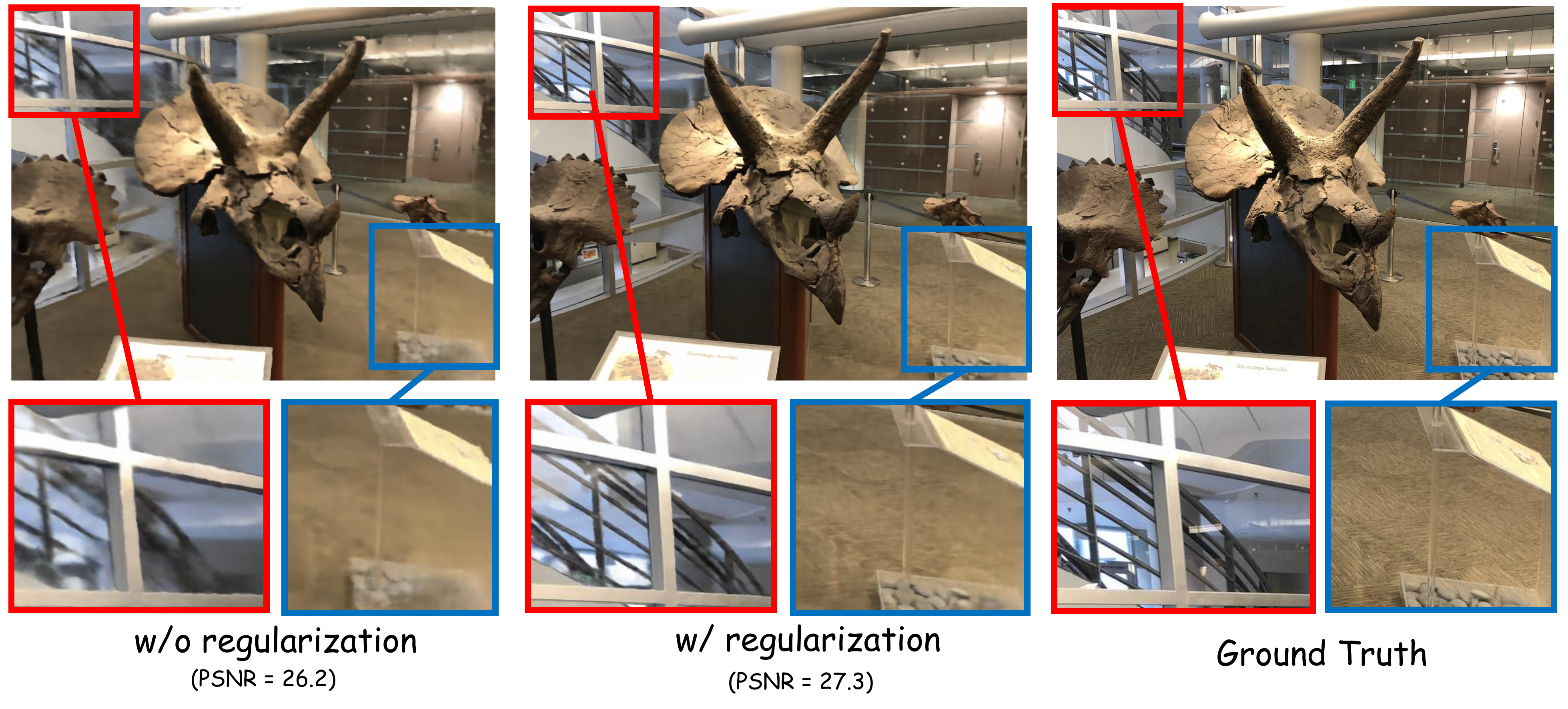}
    % \vskip -0.15in
    \caption{\textbf{Qualitative ablation on regularization term}. The regularization term leads to more apparent synthesis results and greatly alleviates the blurs on the object surfaces. Quantitatively, regularization boosts performance by 1.1 PSNR on this scene}
    \label{figure4}
% \vskip -0.15in
\end{figure}

To optimize our radiance field, we first assume that each location in the scene is at most sampled once in a training batch. We believe the hypothesis is reasonable as the intersection of two rays rarely happens in a 3D scene, let alone sampling at the same position in the same batch. Therefore, the distributions of rendered rays are assumed to be independent. In this way, we can optimize the model by minimizing the negative log-likelihood of rays $\{r_{i=1}^N\}$ from a batch $\mathcal{B}$:

\begin{equation}
\setlength{\abovedisplayskip}{1ex}
    \underset{\theta}{\text{min}} \ -\!\text{log}p_{\theta}(\mathcal{B})\! =\!-\frac{1}{N}\!\sum_{i=1}^N\text{log}\ p_{\theta}(C(\text{r}_i)) =\!\frac{1}{N}\!\sum_{i=1}^N\frac{\|C(\text{r}_i)\!-\!\bar{C}(\text{r}_i)\|_2^2}{2\bar{\beta}^2(\text{r}_i)}\! +\! \frac{\text{log}\ \bar{\beta}^2(\text{r}_i)}{2}.
\setlength{\belowdisplayskip}{1ex}
\end{equation}

However, simply minimizing the above objective function leads to a sub-optimal solution where the weights $\alpha_i$ for different samples in a ray are driven closer. This results in an unexpectedly large fraction of non-zero density in the whole scene, causing blurs on the object's surface, as depicted in Figure \ref{figure4}. Therefore, we add an additional regularization term to force sparser volume density, and the loss function is formulated as:
\begin{equation}
    \mathcal{L}^{uct}\! = \!\frac{1}{N}\!\sum_{i=1}^N \!\left( \frac{\|C(\text{r}_i)\!-\!\bar{C}(\text{r}_i)\|_2^2}{2\bar{\beta}^2(\text{r}_i)}\! +\! \frac{\text{log}\ \bar{\beta}^2(\text{r}_i)}{2}\! +\! \frac{\lambda}{N_s}\! \sum_{j=1}^{N_s}\! \sigma_i(\text{r}_i(t_j)) \right ),
\end{equation}
where $\lambda$ is a hyper-parameter that controls the regularization strength.

We follow the original NeRF framework and optimize two parallel networks. To ease the difficulty of optimization, we only adopt the uncertainty branch in the fine model and keep the coarse model the same as vanilla. The final loss function is then:
\begin{equation}
\setlength{\abovedisplayskip}{1ex}
     \mathcal{L}^{uct}(C(\text{r}), \bar{C}^f(\text{r}))+\frac{1}{N}\sum_{i=1}^N\|C(\text{r}_i)-\hat{C}^c(\text{r}_i)\|_2^2.
\setlength{\belowdisplayskip}{1ex}
\end{equation}

By learning a neural radiance field as Gaussian distributions, we not only produce reasonable predictions in uncertain areas but also present an interpretation of how NeRF model understands the scene. On the one hand, uncertainty can be viewed as a perception of noises, which may also reflect the degree of risk in real-world scenarios, e.g., robotic navigation. On the other hand, this can further serve as a vital criterion in the following active learning framework.

\begin{figure*}[t]
    \centering
    \includegraphics[width=0.95\linewidth]{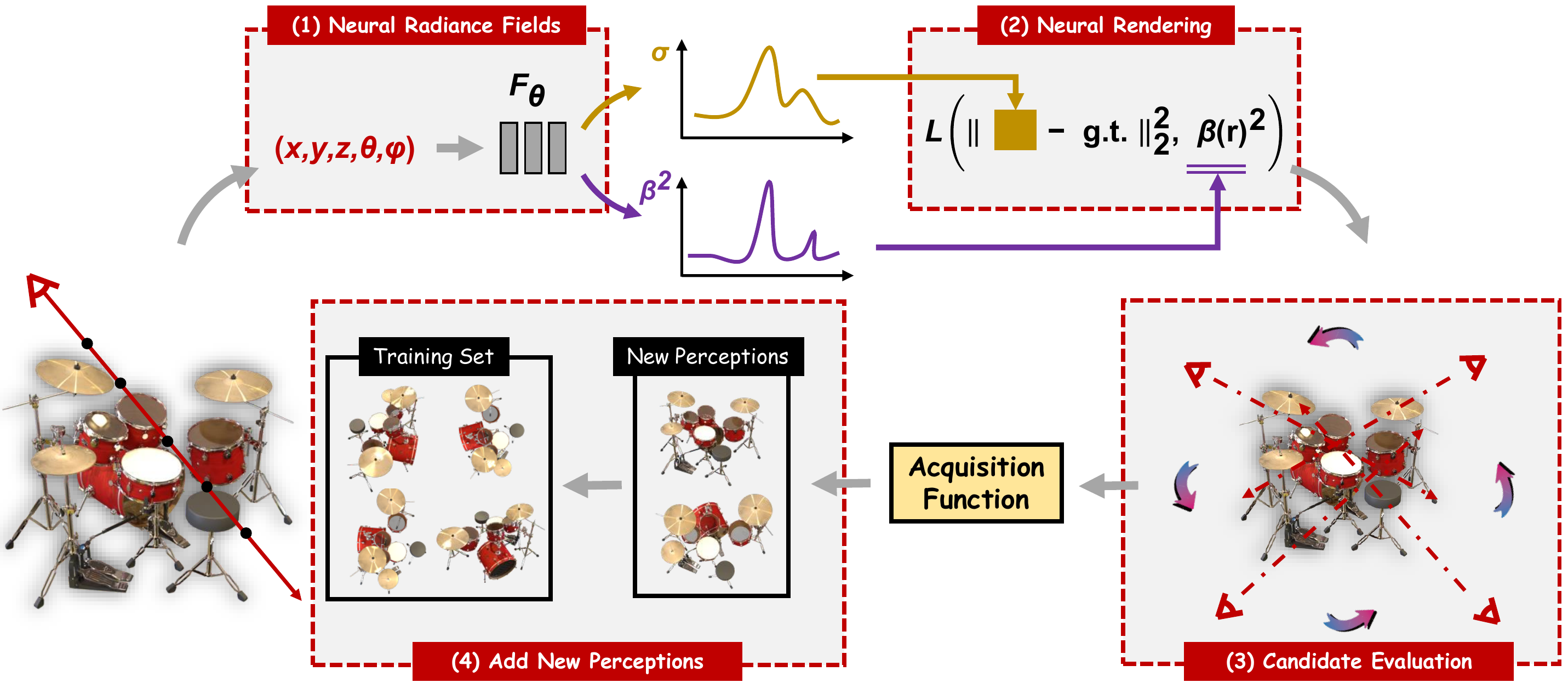}
    \caption{\textbf{The ActiveNeRF pipeline} consists of 4 steps. First, the initial observation is used to train an ActiveNeRF model (Sec.~\ref{ppdistr}). This model is then used to render novel views, from which the new viewpoint (that most reduces uncertainty) is estimated (Sec.~\ref{acq} and \ref{optinf}). Finally, a new perception is captured and added to the training set}
    \label{figure3}
\end{figure*}

\section{ActiveNeRF}
Although several works have attempted to model well-generalized NeRF under a limited training budget, the upper bound of their performances is highly restricted due to the inherent blind spot in the observations. For example, when modeling a car, if the right side of the car is never observed during training, the radiance field in this region would be under-optimized, making it almost impossible to render photo-realistic images.

Different from previous works, we target improving the upper bound of model performances. Inspired by the insights from active learning, we present a novel learning framework named ActiveNeRF and try to supplement the training sample in the most efficient manner, as illustrated in Figure \ref{figure3}. We first introduce how to evaluate the effect of new inputs based on the uncertainty estimation and show two approaches for the framework to incorporate with new inputs. 

\subsection{Prior and Posterior Distribution}
\label{ppdistr}
Estimating the influence of new data without its actual observation is a non-trivial problem. Nevertheless, modeling the radiance field as Gaussian distribution makes the evaluation more tractable, where we can estimate the posterior distribution of the radiance field based on the Bayesian rule.

Let $D_1$ denote the existing training set and $F_{\theta}$ denote the trained NeRF model given $D_1$. For simplicity, we first consider the influence of a single ray $\text{r}_2$ from the new input $D_2$. Thus, for the $k_{th}$ sampled location $\text{r}_2(t_k)$, its prior distribution is formulated as:
\begin{equation}
\setlength{\abovedisplayskip}{1ex}
\label{prior_distribution}
    P^{(\text{pri)}} = P(c(\text{r}_2(t_k))|D_1) \sim \mathcal{N}(\bar{c}(\text{r}_2(t_k)), \bar{\beta}^2(\text{r}_2(t_k))).
\setlength{\belowdisplayskip}{1ex}
\end{equation}
Following the sequential Bayesian formulation, the posterior distribution can then be derived as:
\begin{equation}
\setlength{\abovedisplayskip}{1ex}
\label{post_distribution}
    P^{(\text{post})} = P(c(\text{r}_2(t_k))|D_1,\text{r}_2) = \frac{p(C(\text{r}_2)|c(\text{r}_2(t_k)))p(c(\text{r}_2(t_k))|D_1)}{\int p(C(\text{r}_2)|c(\text{r}_2(t_k)))p(c(\text{r}_2(t_k))|D_1)dc(\text{r}_2(t_k))}.
\setlength{\belowdisplayskip}{1ex}
\end{equation}
As derived in Sec. \ref{uncert_est}, rendered color of rays follows the Gaussian distribution:
\begin{equation}
\setlength{\abovedisplayskip}{1ex}
    p(C(\text{r}_2)|c(\text{r}_2(t_k))) \sim \mathcal{N}(\bar{C}(\text{r}_2), \bar{\beta}^2(\text{r}_2)) \sim \mathcal{N}(\sum_{i=1}^{N_s}\alpha_i\bar{c}(\text{r}_2(t_i)), \bar{\beta}^2(\text{r}_2)).
\setlength{\belowdisplayskip}{1ex}
\end{equation}
As other sampled locations in $\text{r}_2$ are independent with $\text{r}_2(t_k)$, we can represent the unrelated part in the mean as a constant $b(t_k)$ and the distribution can be simplified as:
\begin{equation}
\setlength{\abovedisplayskip}{1ex}
\label{data_distribution}
    p(C(\text{r}_2)|c(\text{r}_2(t_k))) \sim \mathcal{N}(\alpha_k\bar{c}(\text{r}_2(t_k))\!+\!b(t_k), \bar{\beta}^2(\text{r}_2)).
\setlength{\belowdisplayskip}{1ex}
\end{equation}
Finally, by substituting terms in Eq.(\ref{post_distribution}) with Eq.(\ref{prior_distribution}) and Eq.(\ref{data_distribution}), the posterior distribution is formulated as:
\begin{equation}
\setlength{\abovedisplayskip}{1ex}
\label{post}
    P^{(\text{post})} \sim \mathcal{N}\left(  \gamma \frac{C(\text{r}_2)\!-\!b(t_k)}{\alpha_k}\!+\!(1\!-\!\gamma)\bar{c}(\text{r}_2(t_k)), \frac{\bar{\beta}^2(\text{r}_2(t_k))\bar{\beta}^2(\text{r}_2)}{\alpha^2_k\bar{\beta}^2(\text{r}_2(t_k))+\bar{\beta}^2(\text{r}_2)} \right),
\setlength{\belowdisplayskip}{1ex}
\end{equation}
\begin{equation}
\setlength{\abovedisplayskip}{1ex}
    \text{with} \ \ \gamma = \frac{\alpha_k^2\bar{\beta}^2(\text{r}_2(t_k))}{\alpha_k^2\bar{\beta}(\text{r}_2(t_k))+\bar{\beta}^2(\text{r}_2)}.
\setlength{\belowdisplayskip}{1ex}
\end{equation}
Please refer to Appendix A for details.

\subsection{Acquisition Function}
\label{acq}
With the posterior distribution formulated by the Bayesian rule, we quantitatively analyze the influence on the radiance field given a new input ray. As shown in Eq.(\ref{post}), although the mean of the posterior distribution is unavailable due to the unknown of $C(\text{r}_2)$, the variance is independent of the ground truth value and therefore can be \textit{precisely computed} based on the current model $F_{\theta}$. Additionally, it is worth noting that the variance of the posterior distribution of a newly observed location $\text{r}_2(t_k)$ is consistently smaller than its prior distribution:
\begin{align}
    \text{Var}^{(\text{post})}(\text{r}_2(t_k))&\!=\!\frac{\bar{\beta}^2(\text{r}_2(t_k))\bar{\beta}^2(\text{r}_2)}{\alpha^2_k\bar{\beta}^2(\text{r}_2(t_k))+\bar{\beta}^2(\text{r}_2)} \nonumber \\ 
    &\!=\! (\frac{1}{\bar{\beta}^2(\text{r}_2(t_k))}+\frac{\alpha_k^2}{\bar{\beta}^2(\text{r}_2)})^{-1} \!<\! \bar{\beta}^2(\text{r}_2(t_k))\!=\!\text{Var}^{(\text{pri})}(\text{r}_2(t_k)).
\end{align}
This further proves that new observations can genuinely reduce the uncertainty of the radiance field. On this basis, we consider the reduction of variance as the estimation of information gain of $\text{r}_2(t_k)$ from the new ray $\text{r}_2$:
\begin{equation}
\setlength{\abovedisplayskip}{1ex}
    \text{Var}^{(\text{pri})}(\text{r}_2(t_k))-\text{Var}^{(\text{post})}(\text{r}_2(t_k)).
\setlength{\belowdisplayskip}{1ex}
\end{equation}
For a given image with resolution $H,W$, we can sample $N\!=\!H\!\times \!W$ independent rays, with $N_s$ sampled locations from each ray. Therefore, we add up the reduction of variance from all these locations and define the acquisition function as:
\begin{equation}
\setlength{\abovedisplayskip}{1ex}
    \mathcal{A}(D_2) = \sum_{\text{r}_i \in D_2} \sum_{j=1}^{N_s} \left( \text{Var}^{(\text{pri})}(\text{r}_i(t_j))-\text{Var}^{(\text{post})}(\text{r}_i(t_j)) \right).
\setlength{\belowdisplayskip}{1ex}
\end{equation}
Similar derivation is also applicable with multiple input images, where the variance of posterior uncertainty is formulated as:
\begin{equation}
\setlength{\abovedisplayskip}{1ex}
    \text{Var}^{(\text{post})}(\text{x}) = (\frac{1}{\bar{\beta}^2(\text{x})}+\sum_{i}\frac{\alpha^2_{k_i}}{\bar{\beta}^2(\text{r}_i(t_{k_i}))})^{-1},
\setlength{\belowdisplayskip}{1ex}
\end{equation}
where $\text{r}_i$ denotes ray from different images, and $\text{x}=\text{r}_i(t_{k_i}), \forall i$. Please refer to Appendix B for details.

In practical implementation, we first sample candidate views from a spherical space, and choose the top-k candidates that score highest in the acquisition function as the supplementary of the current training set. In this way, the captured new inputs bring the most information gain and promote the performance of the current model with the highest efficiency.

Besides, a quality-efficiency trade-off can also be achieved by evaluating new inputs with lower resolution. For example, instead of using full image size $H\!\times \!W$ as new rays, we can sample $H/r\!\times \!W/r$ rays to approximate the influence of the whole image with only $1/r^2$ time consumption.

\subsection{Optimization and Inference}
\label{optinf}
With the newly captured samples chosen by the acquisition function, we provide two approaches to incorporate the current NeRF model with additional inputs.

\noindent
\textbf{Bayesian Estimation.} With the ground-truth value $C(\text{r})$ from the new inputs, we can practically compute the posterior distribution of the locations in the scene by leveraging Eq.(\ref{post}). Among these, the mean of distribution becomes the Bayesian estimation of emitted radiance value, and can be adopted in the rendering process. At inference time, we only need to substitute the prior color with the posterior Bayesian estimation:
\begin{equation}
\setlength{\abovedisplayskip}{1ex}
\bar{c}(\text{r}(t_k)) \Rightarrow \gamma \frac{C(\text{r})\!-\!b(t_k)}{\alpha_k}\!+\!(1\!-\!\gamma)\bar{c}(\text{r}(t_k)),
\setlength{\belowdisplayskip}{1ex}
\end{equation}
while others remain unchanged.

One of the advantages of using Bayesian estimation is that we avoid the collateral training procedure. If we consider an edge device, \textit{e.g.,} a robot, the training-free scheme allows the agent to perform offline inference instantly, which is more friendly in resource-constrained scenarios.

\noindent
\textbf{Continuous Learning} can also be considered if time and computation resources are not the bottlenecks.
The captured inputs can be added to the training set and tune the model on the basis of the current one. We can further control the fraction of training rays from new images, forcing the model to optimize in the newly observed regions. 

The two approaches can both promote the quality of the neural radiance field, and naturally achieve a trade-off between efficiency and synthesis quality.
\begin{table}[t]
\begin{center}
\tablestyle{5pt}{0.9}
{
\caption{\textbf{Quantitative results in Fixed Training Set setting}: ActiveNeRF performs superior to or on par with the original NeRF in all settings. In particular, note our model performs significantly better than NeRF in low-shot settings. We report PSNR/SSIM (higher is better) and LPIPS (lower is better)}
\label{tab1}
\begin{tabular}{c|ccc|ccc}
 & \multicolumn{3}{c}{\textbf{(a) Synthetic Scenes}} & \multicolumn{3}{c}{\textbf{(b) Realistic Scenes}}\\
 Method & PSNR$\uparrow$ & SSIM$\uparrow$ & LPIPS$\downarrow$ & PSNR$\uparrow$ & SSIM$\uparrow$ & LPIPS$\downarrow$\\
\shline
\multicolumn{7}{l}{\emph{Setting I, training with all images}} \\
\hline
SRN & 22.26 & 0.846 & 0.170 & 22.84 & 0.668 & 0.378\\
LLFF & 24.88 & 0.911 & 0.114 & 24.13 & 0.798 & \textbf{0.212}\\
NeRF & \textbf{31.01} & 0.947 & 0.081 & \textbf{26.50} & 0.811 & 0.250\\
IBRNet & 25.62 & 0.939 & 0.110 & - & - & - \\
MSVNeRF & 27.07 & 0.931 & 0.168 & - & - & - \\
\textbf{Ours} & 30.45 & \textbf{0.954} & \textbf{0.072} & 25.96 & \textbf{0.835} & \textbf{0.213}\\
\hline
\multicolumn{7}{l}{\emph{Setting II, training with 10 images}} \\
\hline
NeRF & 28.04 & 0.866 & 0.134 & 23.36 & 0.791 & 0.280\\
DietNeRF & 28.42 & 0.891 & \textbf{0.087} & - & - & - \\
\textbf{Ours} & \textbf{28.51} & \textbf{0.932} & 0.090 & \textbf{23.96} & \textbf{0.803} & \textbf{0.260}\\
\hline
\multicolumn{7}{l}{\emph{Setting III, training with 5 images}} \\
\hline
 NeRF & 21.14 & 0.835 & 0.192 & 21.67 & 0.689 & 0.350 \\
 \textbf{Ours} & \textbf{23.23} & \textbf{0.866} & \textbf{0.185} & \textbf{22.03} & \textbf{0.712} & \textbf{0.292}\\
\hline
\end{tabular}}
\end{center}
\end{table}

\section{Experiments}
\subsection{Experimental setup}
\noindent
\textbf{Datasets.} We extensively demonstrate our approach in two benchmarks, including LLFF \cite{mildenhall2019local} and NeRF \cite{mildenhall2020nerf} datasets. LLFF is a real-world dataset consisting of 8 complex scenes captured with a cellphone. Each scene contains 20-62 images with $1008\!\times \!756$ resolution, where $1/8$ images are reserved for the test. NeRF dataset contains 8 synthetic objects with complicated geometry and realistic non-Lambertian materials. Each scene has 100 views for training and 200 for the test, and all the images are at $800\!\times \!800$ resolution. See detailed training configurations in the Appendix.

\noindent
\textbf{Metrics.} We report the image quality metrics PSNR and SSIM for evaluations. We also include LPIPS \cite{zhang2018unreasonable}, which more accurately reflects human perception.

\begin{figure}[t]
    \centering
    \includegraphics[width=0.85\linewidth]{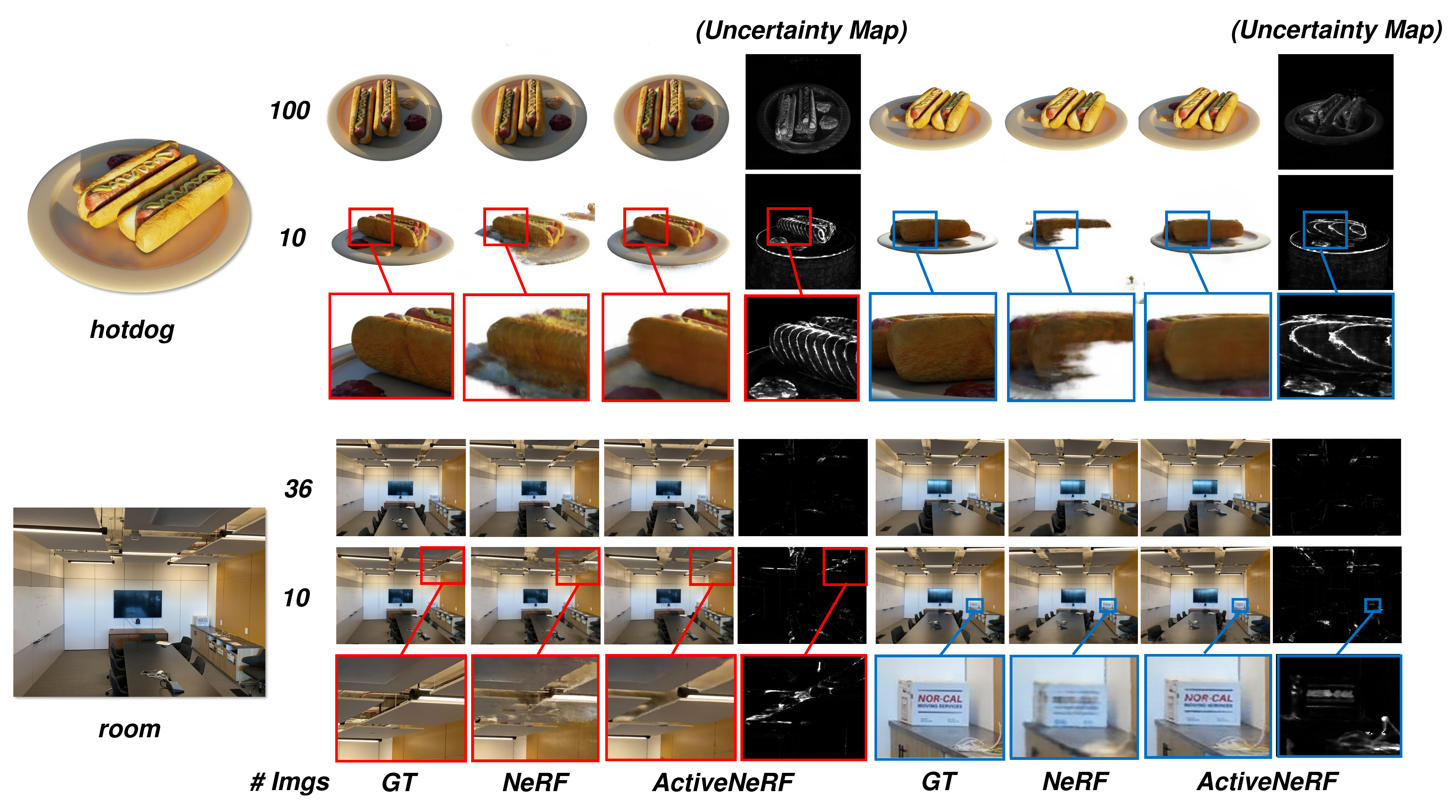}
    \caption{\textbf{Qualitative results on synthetic and realistic scenes} with different fractions of training samples. Several observations can be made: First, ActiveNeRF performs significantly better than NeRF in the low-shot setting (\textit{e.g.,} See Ln. 2 and 3). Second, the uncertainty correctly reduces when more data is used (See Col. \textit{Uncertainty Map}). Finally, ActiveNeRF and NeRF obtain similar qualitative performance when all images are used (See Ln. 1 and 4), suggesting modeling uncertainty has no negative impact on the quality of view synthesis}
    \label{main1}
\end{figure}

\begin{table}[t]
\begin{center}
\tablestyle{2.7pt}{0.95}
{
\caption{\textbf{Quantitative results in Active Learning settings:} \textbf{BE:} Bayesian estimation; \textbf{CL:} Continuous Learning; \textbf{Setting I:} 4 initial observations and 4 extra observations obtained at 40K, 80K, 120K, and 160K iterations. \textbf{Setting II:} 2 initial observations and 2 extra observations are obtained at 40K, 80K, 120K, and 160K iterations. \textbf{NeRF$^\dagger$}: NeRF performance from fixed training set setting. This setting measures NeRF's upper-bound performance by removing the difficulties introduced by continuous learning. Overall, ActiveNeRF outperforms baseline methods; several metrics could even match non-CL performance. We also report the total time consumption (training + inference time) of different approaches in the \textbf{Time} column, where \textbf{ActiveNeRF-BE} only consume training time at first 40K iterations and inference time at later stages}
\label{tab2}
\begin{tabular}{c|cccc|ccc}
 & \multicolumn{3}{c}{\textbf{(a) Synthetic Scenes}} & \multicolumn{3}{c}{\textbf{(b) Realistic Scenes}}\\
 Method & Time & PSNR$\uparrow$ & SSIM$\uparrow$ & LPIPS$\downarrow$ & PSNR$\uparrow$ & SSIM$\uparrow$ & LPIPS$\downarrow$\\
\shline
\multicolumn{8}{l}{\emph{Setting I, 20 total observations:}} \\
\hline
 NeRF+Rand & 2.0h & 24.25 & 0.734 & 0.207 & 20.65 & 0.532 & 0.312\\
 NeRF+FVS & 2.0h & 26.00 & 0.812 & 0.144 & 22.41 & 0.710 & 0.299\\
 \textbf{ActiveNeRF-BE} & 30min & 25.67 & 0.778 & 0.169 & 21.86 & 0.644 & 0.303\\
 \textbf{ActiveNeRF-CL} & 2.2h & \textbf{26.24} & \textbf{0.856} & \textbf{0.124} & \textbf{23.12} & \textbf{0.765} & \textbf{0.292}\\
 \gray{NeRF$^\dagger$} & 2.0h & \gray{28.04} & \gray{0.910} & \gray{0.134} & \gray{23.36} & \gray{0.791} & \gray{0.280}\\
 \hline
\multicolumn{7}{l}{\emph{Setting II, 10 total observations:}} \\
\hline
 NeRF+Rand & 1.0h & 18.36 & 0.642 & 0.251 & 18.49 & 0.478 & 0.355\\
 NeRF+FVS & 1.0h & 19.24 & 0.735 & 0.227 & 20.02 & 0.633 & 0.344\\
 \textbf{ActiveNeRF-BE} & 16min & 18.25 & 0.611 & 0.256 & 18.67 & 0.451 & 0.367\\
 \textbf{ActiveNeRF-CL} & 1.1h & \textbf{20.01} & \textbf{0.832} & \textbf{0.204} & \textbf{20.14} & \textbf{0.664} & \textbf{0.325}\\
 \gray{NeRF$^\dagger$} & 1.0h & \gray{21.14} & \gray{0.835} & \gray{0.192} & \gray{21.67} & \gray{0.689} & \gray{0.350}\\
\hline
\end{tabular}}
\end{center}
\end{table}

\begin{figure}[t]
    \centering
    \includegraphics[width=.85\linewidth]{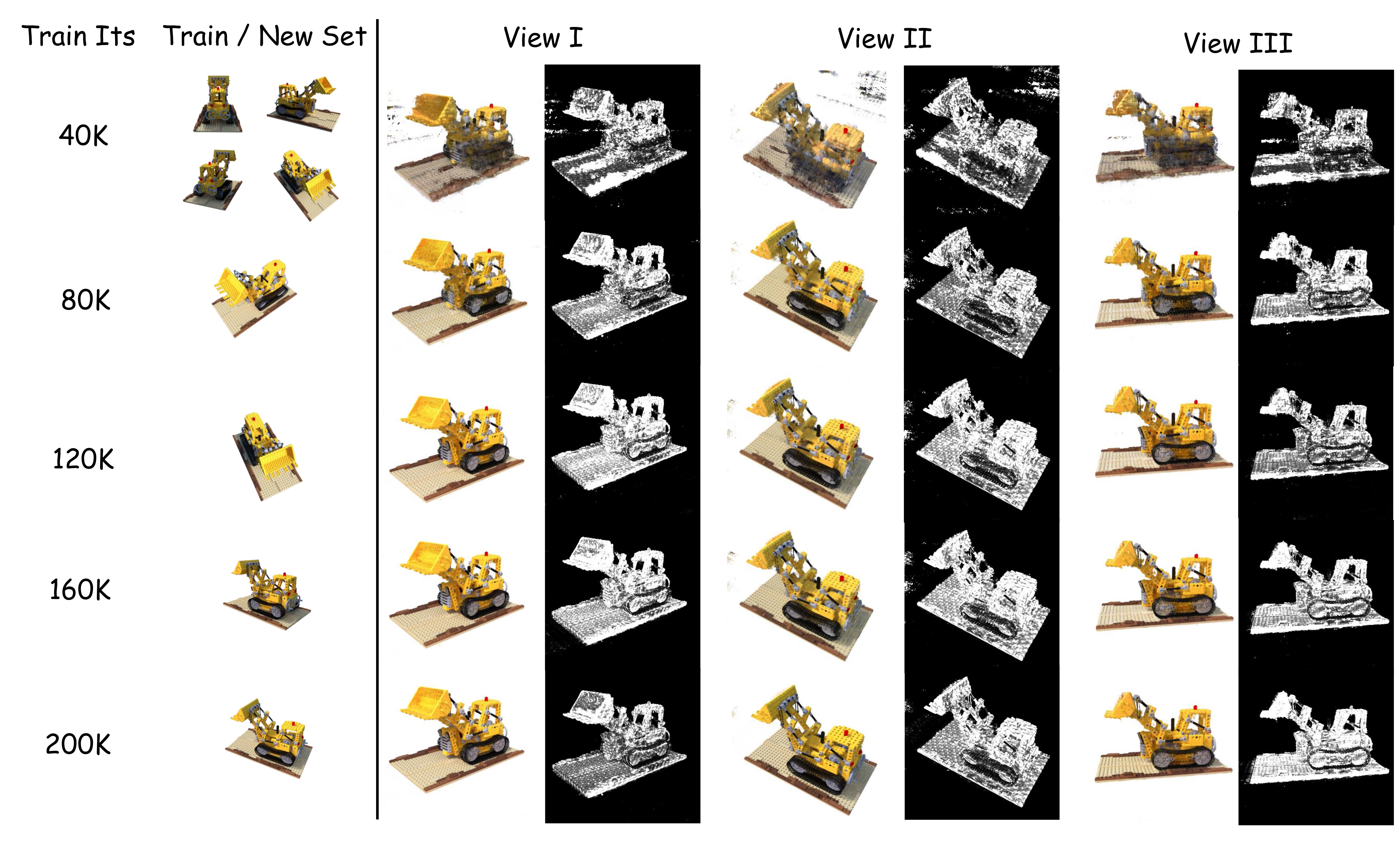}
    \caption{\textbf{Qualitative results of ActiveNeRF with four active iterations.} We capture new perceptions every 40K iterations. Improved synthesis quality can be seen in unobserved regions}
    \label{main2}
\end{figure}

\subsection{Results}
\subsubsection{Uncertainty Estimation.}
We first evaluate the effectiveness of the proposed uncertainty estimation with different fractions of input samples. We compare with several competitive baselines, including Neural Radiance Fields (NeRF) \cite{wang2021nerf}, Local Light Field Fusion (LLFF) \cite{mildenhall2019local}, and Scene Representation Networks (SRN) \cite{sitzmann2019scene}. We also compare with three competitive baselines, including IBRNet \cite{wang2021ibrnet}, MSVNeRF \cite{chen2021mvsnerf} and DietNeRF \cite{jain2021putting}.

We show the performance of our proposed approach with a different number of training data over baseline approaches in Table \ref{tab1}. It can be seen that NeRF with uncertainty performs on par or slightly better than baseline models, showing that modeling uncertainty does not affect the quality of synthesizing novel views. When it comes to limited training samples, our model shows consistently better results. For example, in the synthetic dataset, NeRF with $10\%$ training data fails to generalize well to some views, while our model can still provide reasonable predictions. The gap is more distinct on the perceptual loss, \textit{e.g.,} LPIPS, showing that our model can also render high-frequency textures with limited training data. Compared to DietNeRF, our model achieves better performances on two criteria and is competitive on the third. However, ours do not require additional pretrained model (\textit{e.g.,} CLIP for DietNeRF) and can be used in the following active learning framework. Qualitative results are shown in Figure \ref{main1}.

\subsubsection{ActiveNeRF.}
We validate the performance of our proposed framework, ActiveNeRF, and compare it with two heuristic approaches. As an approximation, we hold out a large fraction of images in the training set and use these images as candidate samples. For baselines, we denote \textit{NeRF+Random} as randomly capturing new images in the candidates. \textit{NeRF+FVS (furthest view sampling)} corresponds to finding the candidates with the most distanced camera position compared with the current training set. We empirically adjust the number of the initial training set and captured samples during the training procedure.

We first show the results with continuous learning scheme, where the time and computation resources are considered sufficient. The comparison results are shown in Table \ref{tab2} and Figure \ref{main2}. We can easily see that ActiveNeRF captures the most informative inputs comparing with heuristic approaches, which contributes most to synthesizing views from less observed regions. The additional training cost for ActiveNeRF is also comparably minor (2.2h vs. 2h).

We further validate the model performances with Bayesian estimation. As shown in Table \ref{tab2}, $75\%$ of the time consumption can be saved. Although showing inferior performance to continuous learning, the model with Bayesian estimation still synthesize reasonable images and is even competitive with heuristic approaches under continuous learning scheme.
\section{Conclusion}
In this paper, we present a flexible learning framework, that supplements the existing training set with newly captured samples based on an active learning scheme. We first incorporate uncertainty estimation into a NeRF model and evaluate the reduction of uncertainty in the scene given new inputs. By selecting the samples that bring the most information gain, the quality of novel view synthesis can be promoted with minimal additional resources. Also, our approach can be applied to various NeRF-extension approaches as a plug-in module, and enhance model performances in a resource-efficient manner.

\section*{Acknowledgement}
This work is supported in part by National Key R\&D Program of China (2021ZD0140407), the National Natural Science Foundation of China under Grants 62022048 and THU-Bosch JCML Center Beijing Academy of Artiﬁcial Intelligence.
\section*{A. Derivation of Posterior Distribution (Eq.(17))}
\noindent
Given prior distribution as:
\begin{equation}
\label{prior_distribution}
    P^{(\text{pri)}} = P(c(\text{r}_2(t_k))|D_1) \sim \mathcal{N}(\bar{c}(\text{r}_2(t_k)), \bar{\beta}^2(\text{r}_2(t_k))).
\end{equation}
Given new data distribution as:
\begin{equation}
\label{data_distribution}
    p(C(\text{r}_2)|c(\text{r}_2(t_k))) \sim \mathcal{N}(\alpha_k\bar{c}(\text{r}_2(t_k))\!+\!b(t_k), \bar{\beta}^2(\text{r}_2)).
\end{equation}
The corresponding posterior distribution can be formulated as:
\begin{equation}
\label{post}
    P^{(\text{post})} \sim \mathcal{N}\left(  \gamma \frac{C(\text{r}_2)\!-\!b(t_k)}{\alpha_k}\!+\!(1\!-\!\gamma)\bar{c}(\text{r}_2(t_k)), \frac{\bar{\beta}^2(\text{r}_2(t_k))\bar{\beta}^2(\text{r}_2)}{\alpha^2_k\bar{\beta}^2(\text{r}_2(t_k))+\bar{\beta}^2(\text{r}_2)} \right),
\end{equation}
\begin{equation}
    \text{with} \ \ \gamma = \frac{\alpha_k^2\bar{\beta}^2(\text{r}_2(t_k))}{\alpha_k^2\bar{\beta}(\text{r}_2(t_k))+\bar{\beta}^2(\text{r}_2)}.
\end{equation}

\begin{proof}
\begin{align}
    \label{post_distribution}
    &P^{(\text{post})} = P(c(\text{r}_2(t_k))|D_1,\text{r}_2) = \frac{p(C(\text{r}_2)|c(\text{r}_2(t_k)))p(c(\text{r}_2(t_k))|D_1)}{\int p(C(\text{r}_2)|c(\text{r}_2(t_k)))p(c(\text{r}_2(t_k))|D_1)dc(\text{r}_2(t_k))} \\
    &\propto \text{exp}\left( \!-\!\frac{(C(\text{r}_2)\!-\!\alpha_kc(\text{r}_2(t_k))\!-\!b(t_k))^2}{2\bar{\beta}^2(\text{r}_2)} \right) \text{exp}\left(\!-\!\frac{(c(\text{r}_2(t_k)\!-\!\bar{c}(\text{r}_2(t_k)))^2}{2\bar{\beta}^2(\text{r}_2(t_k))} \right) \\
    &\propto \text{exp}\!\left(\!-\!\frac{1}{2}\frac{\alpha^2_k\bar{\beta}^2(\text{r}_2(t_k))+\bar{\beta}^2(\text{r}_2)}{\bar{\beta}^2(\text{r}_2(t_k))\bar{\beta}^2(\text{r}_2)} (c(\text{r}_2(t_k))\!-\!\gamma \frac{C(\text{r}_2)\!-\!b(t_k)}{\alpha_k}\!-\!(1\!-\!\gamma)\bar{c}(\text{r}_2(t_k)))^2\right).
\end{align}
\end{proof}

\section*{B. Derivation of Posterior Distribution with Multiple New Data (Eq.(22))}
\noindent
For simplicity, we consider the situation with two new inputs, and the derivations for more inputs are similar. We change some symbols in the main paper with short notations. The derivation in the main paper can be applied readily with simple substitutions. For $\text{x}=\text{r}_1(t_1)=\text{r}_2(t_2)$, given prior distribution as:
\begin{equation}
\label{prior_distribution}
    P^{(\text{pri})} = P(c(\text{x})|D_1) \sim \mathcal{N}(\bar{c}(\text{x}), \beta_0^2).
\end{equation}
Given new data distribution as:
\begin{align}
    p(\text{r}_1|\text{x}) \sim \mathcal{N}(\alpha_1 \text{x}+b_1, \beta_1^2), \\
    p(\text{r}_2|\text{x}) \sim \mathcal{N}(\alpha_2 \text{x}+b_2, \beta_2^2).
\end{align}
The corresponding posterior distribution can be formulated as:
\begin{equation}
\label{post}
    P^{(\text{post})} \sim \mathcal{N}\left(  \gamma_1 \frac{C(\text{r}_1)\!-\!b_1}{\alpha_1}\!+\!\gamma_2 \frac{C(\text{r}_2)\!-\!b_2}{\alpha_2}\!+\!\gamma_3 \bar{c}(\text{x}), \gamma_3 \beta_0^2 \right),
\end{equation}
where
\begin{align}
    \gamma_1 = \frac{\alpha_1^2\beta_0^2\beta_2^2}{\alpha_1^2\beta_0^2\beta_2^2 + \alpha_2^2\beta_0^2\beta_1^2 + \beta_1^2\beta_2^2}, \\
    \gamma_2 = \frac{\alpha_2^2\beta_0^2\beta_1^2}{\alpha_1^2\beta_0^2\beta_2^2 + \alpha_2^2\beta_0^2\beta_1^2 + \beta_1^2\beta_2^2}, \\
    \gamma_3 = \frac{\beta_1^2\beta_2^2}{\alpha_1^2\beta_0^2\beta_2^2 + \alpha_2^2\beta_0^2\beta_1^2 + \beta_1^2\beta_2^2}.
\end{align}

\begin{proof}
\begin{align}
    \label{post_distribution}
    &P^{(\text{post})} = P(c(\text{x})|D_1,\text{r}_1,\text{r}_2) = \frac{p(C(\text{r}_1)|c(\text{x}))p(C(\text{r}_2)|c(\text{x}))p(c(\text{x})|D_1)}{\int p(C(\text{r}_1)|c(\text{x}))p(C(\text{r}_2)|c(\text{x}))p(c(\text{x})|D_1)dc(\text{x})} \\
    &\propto \text{exp}\left( \!-\frac{(C(\text{r}_1)\!-\!\alpha_1 c(\text{x})\!-\!b_1)^2}{2\beta_1^2}\!-\!\frac{(C(\text{r}_2)\!-\!\alpha_2 c(\text{x})\!-\!b_2)^2}{2\beta_2^2}\!-\!\frac{(c(\text{x})-\bar{c}(\text{x}))^2}{2\beta_0^2} \right) \\
    &\propto \text{exp}\left(\!-\frac{1}{2}\frac{1}{\gamma_3 \beta_0^2} (c(\text{x})\!-\!\gamma_1 \frac{C(\text{r}_1)\!-\!b_1}{\alpha_1}\!-\!\gamma_2 \frac{C(\text{r}_2)\!-\!b_2}{\alpha_2}\!-\!\gamma_3 \bar{c}(\text{x}))^2\right).
\end{align}
\end{proof}

\begin{figure}[h]
    \centering
    \includegraphics[width=.9\linewidth]{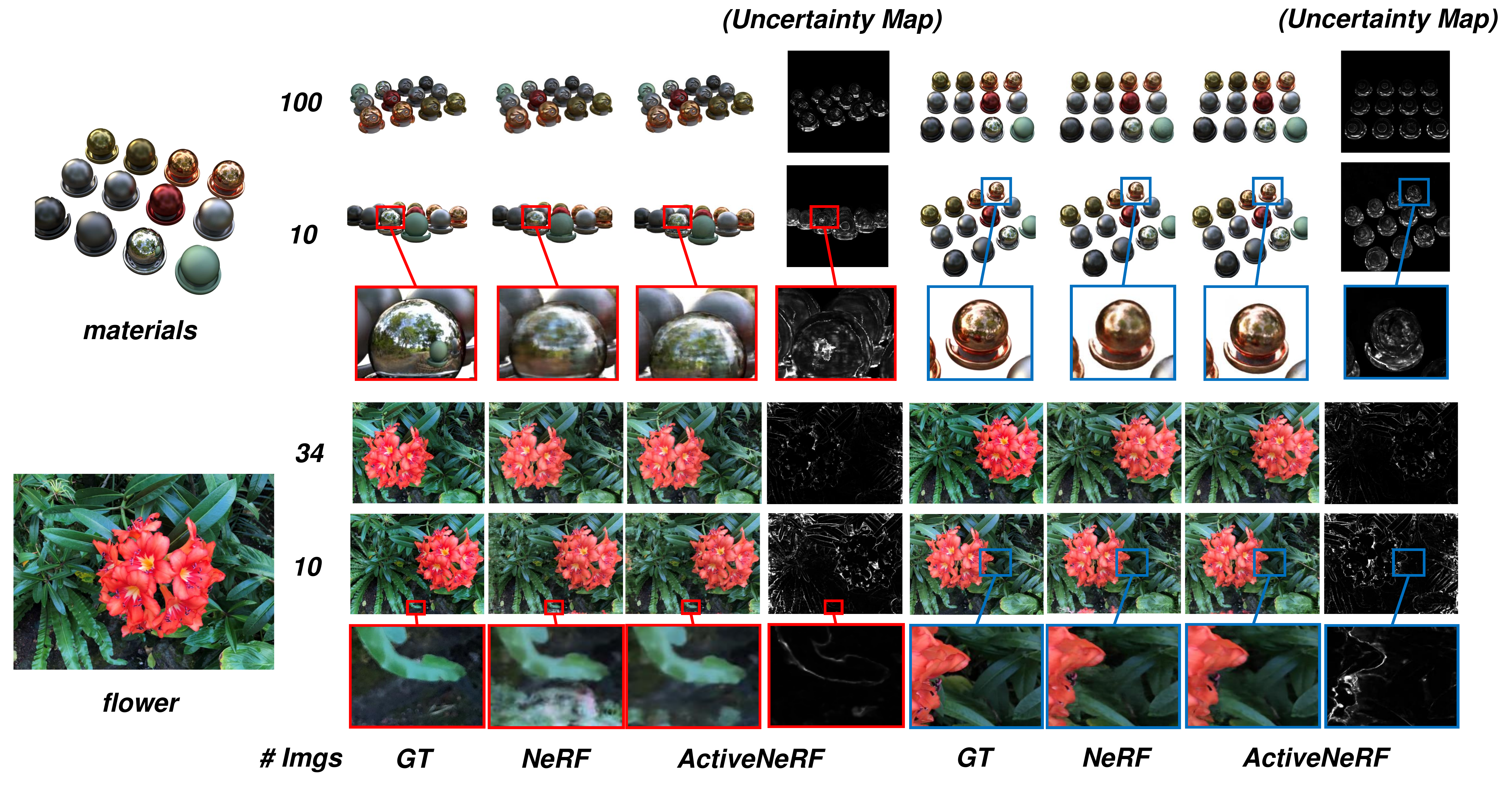}
    \caption{\textbf{Qualitative results on synthetic and realistic scenes} with different fractions of training samples. Several observations can be made: First, ActiveNeRF performs significantly better than NeRF in the low-shot setting (\textit{e.g.,} See Ln. 2 and 3). Also, ActiveNeRF and NeRF obtain similar qualitative performance when all images are used (See Ln. 1 and 4), suggesting modeling uncertainty has no negative impact on the quality of view synthesis.}
    \label{supp1}

\end{figure}
\begin{figure}[h]
    \centering
    \includegraphics[width=.9\linewidth]{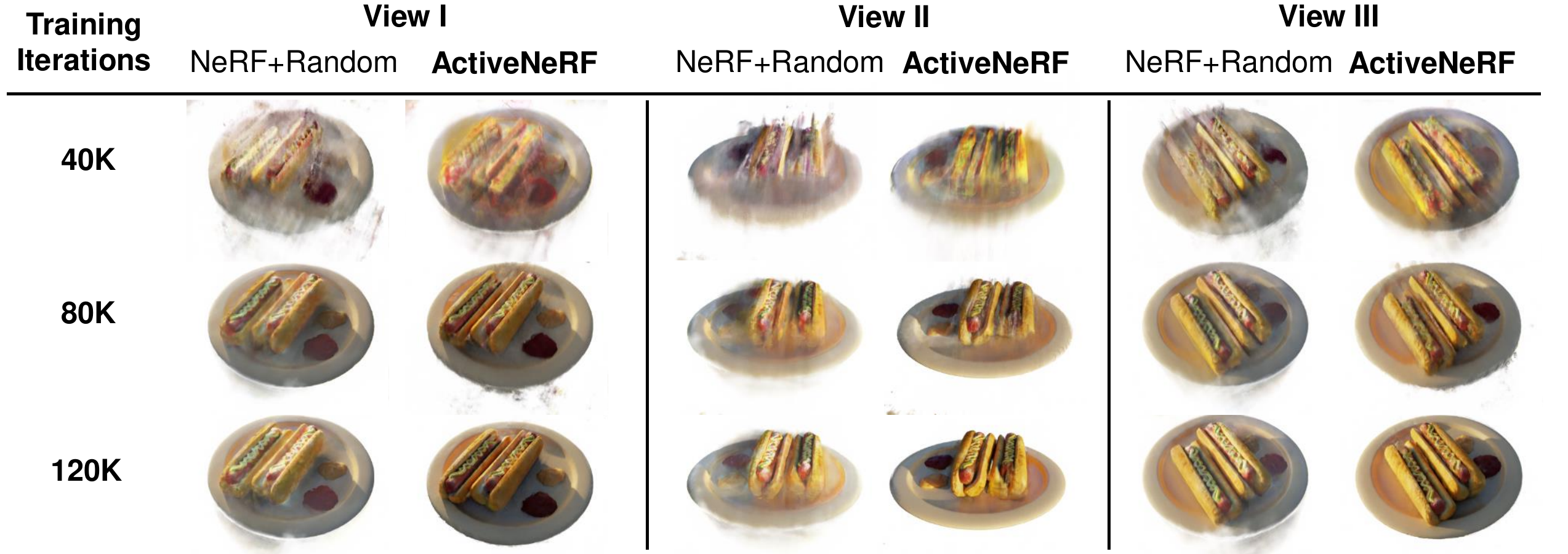}
    \caption{\textbf{Comparison of ActiveNeRF with \textit{NeRF+Random}}. The initial training set has 4 images, and we capture 4 new perceptions every 40K iterations. Our candidate evaluation function is proved to outperform heuristic approaches on the quality of view synthesis.}
    \label{supp2}
\end{figure}

\begin{figure}[h]
    \centering
    \includegraphics[width=.9\linewidth]{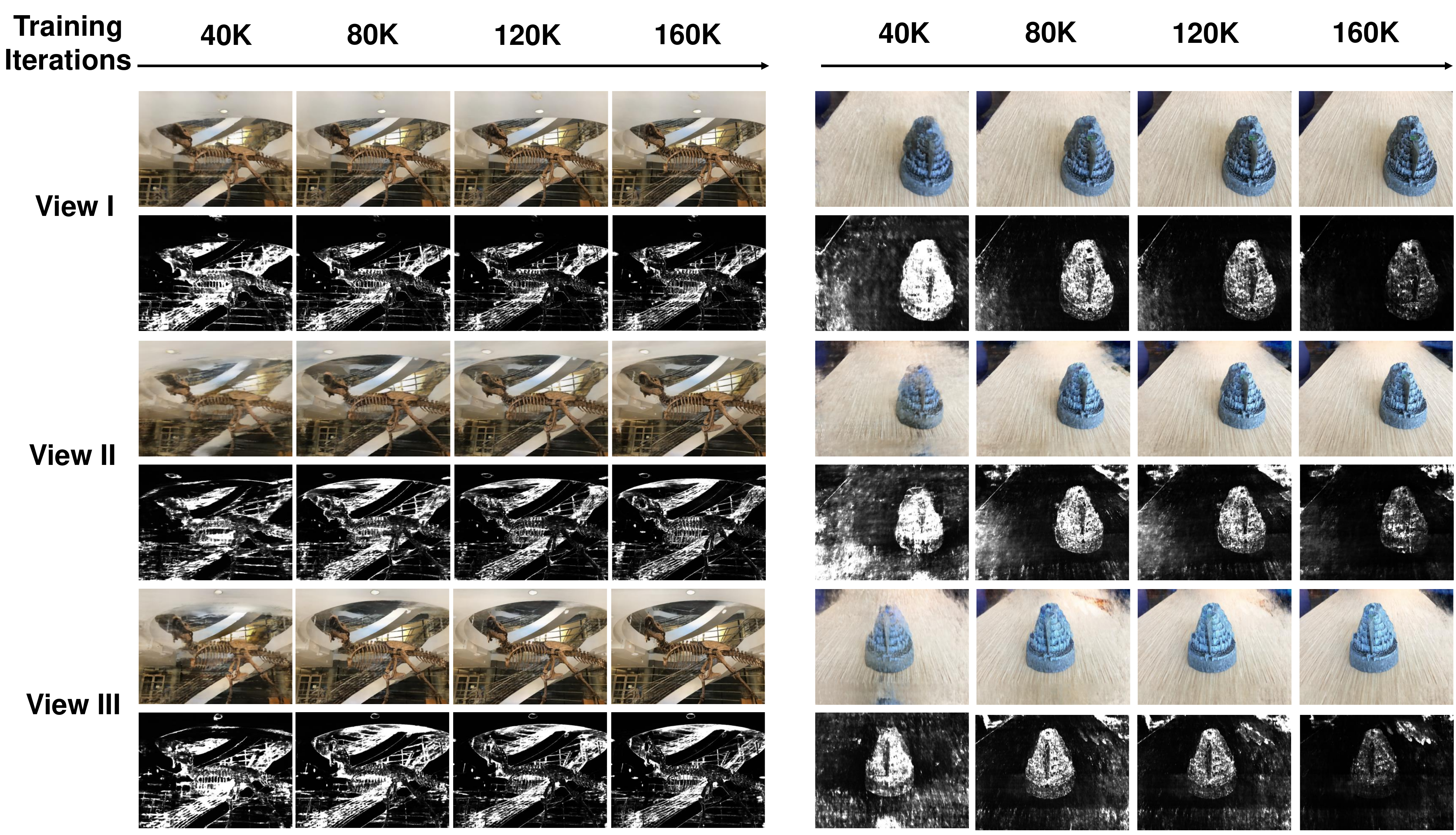}
    \caption{\textbf{Qualitative results of ActiveNeRF with active iterations on realistic scenes.} We capture new perceptions every 40K iterations. Improved synthesis quality can be observed at unobserved regions.}
    \label{supp3}
\end{figure}

\section*{C. Model Architectures}
The detailed model architecture for ActiveNeRF is similar to the original NeRF \cite{mildenhall2020nerf}. We first use the positional encoding function to generate high frequency details. Specifically, the encoding function is formulated as:
\begin{equation}
    \gamma(p)\!=\!(\text{sin}(2^0\pi p),\! \text{cos}(2^0\pi p),\! \cdot\!\cdot\!\cdot\!,\! \text{sin}(2^{L\!-\!1}\pi p),\! \text{cos}(2^{L\!-\!1}\pi p)),
\end{equation}
which separately applied to 3D coordinates $x,y,z$ and Cartesian viewing direction $d_x, d_y, d_z$. We follow the configurations in NeRF and set $L\!=\!10$ for coordinates, and $L\!=\!4$ for directions.

A MLP with 8 fully-connected layers is first adopted to process the encoded 3D coordinates, with residual connection on the $4_{th}$ layer. Then, a single fully connected layer is adopted to predict the volume density $\sigma$. Different from original NeRF, we add an additional fully connected layer with softplus activation function to predict the corresponding variance. The latent feature is then concatenated with encoded viewing directions to produce RGB color.

\section*{D. Training Configurations.}
In our experiments, we follow the settings in NeRF, and sample 64, 128 points for coarse and fine models respectively. We use the Adam optimizer with an initial learning rate at $5e^{-4}$ which decays exponentially to $5e^{-5}$ during optimization. We use a batch size of 1024 rays and train our model on a single RTX2080Ti GPU.

\section*{E. Additional Qualitative Results}
We provide additional visualization results for static and active scenarioes, as shown in Figure \ref{supp1}$\sim$\ref{supp3}.

\section*{F. Ablation Study}
We further evaluate the effectiveness of the candidate evaluation module alone. We substitute the acquisition function with heuristic approaches while keeps the uncertainty module. We denote the baseline approaches as \textit{ActiveNeRF + Random} and \textit{ActiveNeRF + FVS} as introduced in the main paper. The results are shown in Table \ref{tab1} and ActiveNeRF outperforms other approaches consistently.

\begin{table}[h]
\begin{center}
\tablestyle{10pt}{1.07}{
\caption{\textbf{Ablation on acquisition function }: \textit{Setting I} is same as the setting I in the Table 2 of main paper, which includes 4 initial observations and 4 extra observations obtained at 40K, 80K,120K and 160K iterations. } 
\begin{tabular}{c|ccc}
 Sampling Rate (r) & PSNR$\uparrow$ & SSIM$\uparrow$ & LPIPS$\downarrow$\\
\shline
\multicolumn{4}{l}{\emph{Setting I, 20 total observations:}} \\
\hline
 \textbf{ActiveNeRF (Full)} & \textbf{26.24} & \textbf{0.856} & \textbf{0.124}\\
ActiveNeRF+FVS &  26.05 & 0.852 & 0.146\\
ActiveNeRF+Random & 24.77 & 0.801 & 0.188\\
\hline
\end{tabular}}
\end{center}
\label{tab1}
\end{table}

We also investigate the performance of ActiveNeRF when the new perceptions are evaluated in a lower resolution. It can be seen in Table \ref{tab2} that our model can still achieve competitive performances with sampling rate as 5, and reduces the time consumption. When the sampling rate becomes larger than 10, the quality of novel view synthesis is gradually affected.

\begin{table}[h]
\begin{center}
\tablestyle{10pt}{1.07}{
% \small
% \setlength{\tabcolsep}{1.0mm}{
% \renewcommand\arraystretch{1.0}
\caption{\textbf{Ablation on sampling rate}}
\label{tab2}
\begin{tabular}{c|cccc}
% \thickhline
% \multicolumn{5}{c}{\textbf{Ablation Study - Synthetic Scenes}} \\
 Sampling Rate (r) & Time & PSNR$\uparrow$ & SSIM$\uparrow$ & LPIPS$\downarrow$\\
\shline
\multicolumn{5}{l}{\emph{Setting I, 20 total observations:}} \\
\hline
 1 (Full) & 2.20h & \textbf{26.24} & \textbf{0.856} & \textbf{0.124}\\
 5 & 2.12h & 26.12 & 0.855 & 0.124\\
 10 & 2.10h & 25.15 & 0.812 & 0.135\\
 20 & 2.09h & 24.67 & 0.799 & 0.167\\
\hline
\end{tabular}}
\end{center}
\end{table}
\bibliographystyle{splncs04}
\bibliography{egbib}
\end{document}